\documentclass[11pt]{article}

\usepackage[margin=1in]{geometry}
\usepackage[T1]{fontenc}
\usepackage[utf8]{inputenc}
\usepackage{amsmath,amssymb}
\usepackage{graphicx}
\usepackage{url}
\usepackage[authoryear,round]{natbib}
\usepackage{hyperref}

\hypersetup{
  colorlinks=true,
  linkcolor=blue,
  citecolor=blue,
  urlcolor=blue
}

\title{OpenSR-SRGAN: A Flexible Super-Resolution Framework for Multispectral Earth Observation Data}
\author{
Simon Donike\thanks{Corresponding author: \href{mailto:simon.donike@uv.es}{simon.donike@uv.es}. ORCID: \href{https://orcid.org/0000-0002-4440-3835}{0000-0002-4440-3835}.}
\qquad Cesar Aybar\thanks{ORCID: \href{https://orcid.org/0000-0003-2745-9535}{0000-0003-2745-9535}.}
\\[0.75ex]
Julio Contreras\thanks{ORCID: \href{https://orcid.org/0009-0001-5408-7055}{0009-0001-5408-7055}.}
\qquad Luis G\'omez-Chova\thanks{ORCID: \href{https://orcid.org/0000-0003-3924-1269}{0000-0003-3924-1269}.}
\\[0.75ex]
\small Image and Signal Processing Group, University of Valencia, Spain
}
\date{\small Published in \textit{Journal of Open Source Software}, 11(126), 11254 (2026).\\
\href{https://doi.org/10.21105/joss.11254}{doi:10.21105/joss.11254}}

\begin{document}
\maketitle

\begin{abstract}
Satellite images often contain useful measurements beyond visible red, green and blue light, but at a spatial resolution that is too coarse for some applications. Super-resolution methods estimate a finer-resolution image from these coarse observations. OpenSR-SRGAN is a modular, configuration-driven framework for training and benchmarking such models on multispectral Earth-observation data. The software enables users to train and evaluate configurable generator--discriminator architectures on arbitrary sensor band setups using concise YAML configurations, without modifying model code. All components, such as generators, discriminators, loss functions, training schedules, normalizations, and stability heuristics, are exposed through a common YAML interface. OpenSR-SRGAN makes it straightforward to reproduce experiments, compare architectures, and deploy super-resolution pipelines across diverse remote-sensing datasets.

OpenSR-SRGAN supports complete end-to-end workflows with minimal setup: selecting architectures, scaling factors, band combinations, and training strategies entirely from configuration files. Although designed for remote-sensing super-resolution, its tensor-based model components can be adapted to other paired imaging data when suitable preprocessing and validation are provided.
\end{abstract}

\begin{figure}[!ht]
\centering
\IfFileExists{rgb_4x_error.png}{%
  \includegraphics[width=0.80\textwidth]{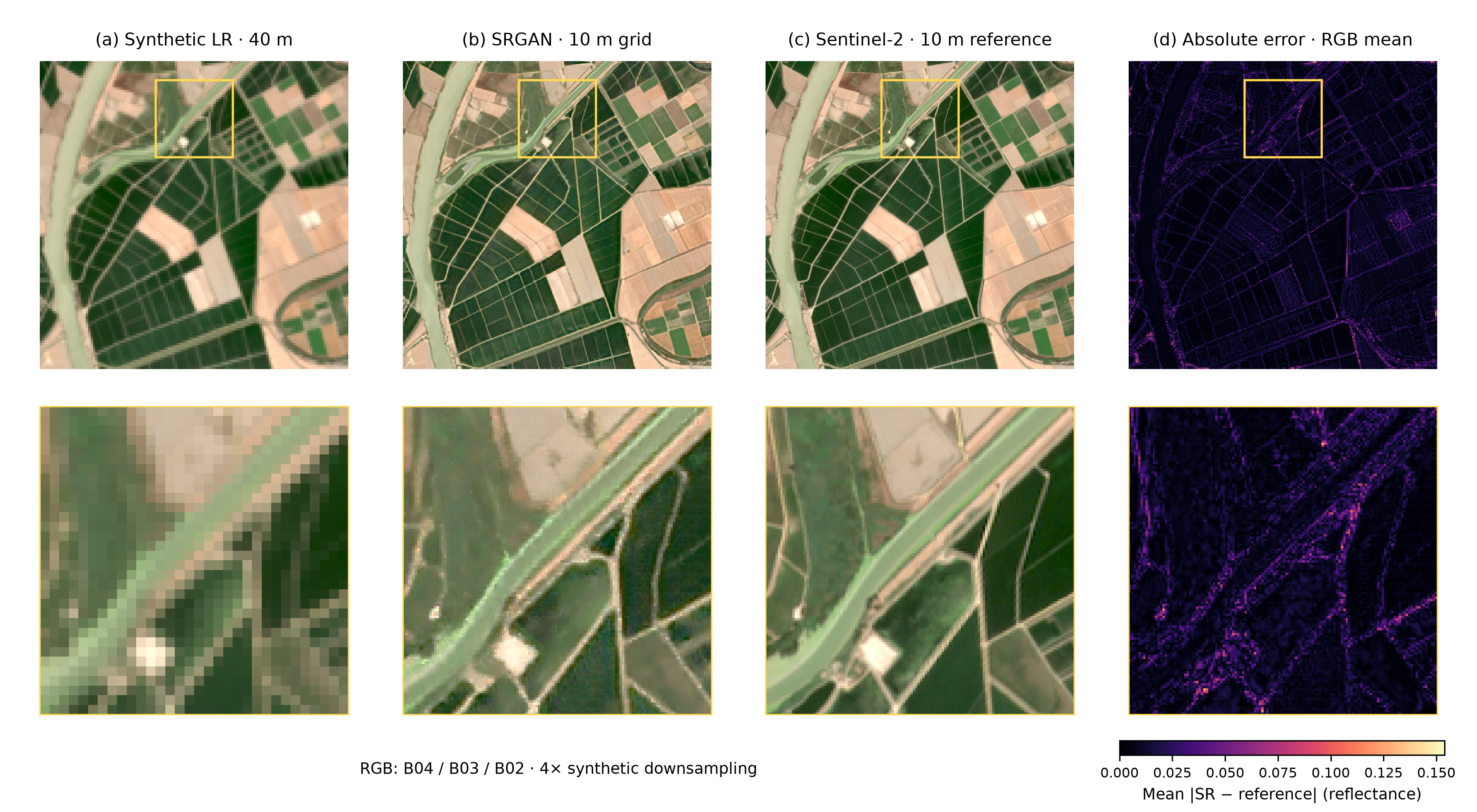}%
}{%
  \fbox{\parbox[c][2.2cm][c]{0.92\textwidth}{\centering
    \textbf{Missing figure asset}\\
    \texttt{rgb\_4x\_error.png}\\
    Add this image before submitting the arXiv source.}}%
}
\caption{Sentinel-2 RGB (B04/B03/B02) comparison for a synthetic 4$\times$ super-resolution task: (a) input downsampled to 40 m; (b) OpenSR-SRGAN output on a 10 m grid; (c) the original 10 m Sentinel-2 reference; and (d) mean absolute SR--reference error across the three RGB bands, computed in reflectance units before display stretching. The bottom row shows matching enlargements of the regions outlined in yellow. The overview and enlarged error maps use the same linear colour scale.}
\label{fig:banner}
\end{figure}

\section{Introduction}

Optical satellite imagery supports many geospatial applications, including agriculture \citep{agriculture}, land-cover mapping \citep{mapping}, ecosystem assessment \citep{ecosysetm} and disaster monitoring \citep{disaster}. Sensors such as Sentinel-2 provide rich multispectral data but at moderate spatial resolution, motivating single-image super-resolution (SISR) to recover finer detail from coarse observations.

Deep learning has driven major progress in SISR, with convolutional models improving fidelity and perceptual quality \citep{dong2015imagesuperresolutionusingdeep,kim2016deeplyrecursiveconvolutionalnetworkimage}. GANs \citep{goodfellow2014generativeadversarialnetworks} introduced adversarial learning for synthesizing realistic high-frequency detail and are widely used in remote sensing \citep{11159252,su2024intriguingpropertycounterfactualexplanation}. SRGAN \citep{ledig2017photo} extended these ideas to super-resolution and has been applied to multispectral data \citep{rs15205062,9787539,10375518,satlassuperres}.

Although diffusion and transformer-based models increasingly define the state of the art \citep{s1,s2,s3}, GAN-based SR methods remain relevant for efficient and deterministic enhancement workflows \citep{g1,satlassuperres}.

\section{Statement of Need}

GANs remain difficult to train \citep{p1,p2,p3}, and these challenges are amplified in remote sensing, where models must handle multispectral inputs, high dynamic-range reflectance values, heterogeneous sensor characteristics, and limited availability of aligned high-resolution reference data. Many research implementations reproduce a single architecture or assume RGB imagery, offering limited flexibility for modifying band configurations, normalization schemes, loss compositions, or training strategies. As a result, researchers who need to adjust models for different sensors (e.g., Sentinel-2, SPOT, Pleiades, or PlanetScope) must often re-engineer core components, modify low-level code, and manually implement stabilization heuristics such as warmup, ramping, or EMA tracking, as well as logging mechanisms. This makes reproducing published experiments or conducting systematic comparisons across architectures labor-intensive, brittle, and inconsistent across studies. These challenges call for a flexible, extensible and configuration-first framework that reduces implementation overhead while enabling systematic experimentation across architectures, loss designs and multispectral sensor configurations. The primary audience is researchers and practitioners developing, comparing, or deploying multispectral super-resolution models for Earth-observation data.

\section{State of the Field}

General-purpose restoration frameworks such as BasicSR\footnote{\url{https://github.com/XPixelGroup/BasicSR}} and MMagic\footnote{\url{https://github.com/open-mmlab/mmagic}} provide configurable implementations of SRGAN, ESRGAN, and other super-resolution models. TorchGeo provides complementary datasets, samplers, and multispectral utilities for geospatial machine learning \citep{stewart2022torchgeo}. The BasicSR package is broad, but does not provide the complete combination of multispectral GAN configurations, reflectance-aware normalization, spectral losses, stabilization schedules, Sentinel-2 presets, and evaluation and deployment interfaces necessary for remote-sensing specific workflows.

OpenSR-SRGAN was developed as a separate focused package because its contribution is this shared Earth-observation experimentation protocol rather than a new generator architecture. Implementing the same functionality as isolated additions to a general restoration framework would still leave researchers responsible for assembling and maintaining the sensor-specific training, evaluation, and large-scene inference workflow. OpenSR-SRGAN instead integrates these choices in one compact interface while reusing established libraries and the wider OpenSR ecosystem.

\section{Software Design}

OpenSR-SRGAN provides a unified, modular and configuration-driven framework for training and evaluating GAN-based super-resolution models for multispectral remote sensing data. All components of an experiment, including architectures, losses, optimizers, data pipelines and training behavior, are defined through concise YAML files, which makes runs repeatable and auditable without modifying source code. The framework supports arbitrary band configurations and adapts naturally to different sensors.

The configuration-first design was chosen to make comparisons auditable: generators, discriminators, losses, and training schedules are instantiated through shared factories. Flat YAML configurations remain easy to archive, while Hydra-based composition supports reusable presets and command-line overrides. This design favors a stable and inspectable experiment schema.

The main features include:
\begin{itemize}
  \item \textbf{Modular GAN framework:} Interchangeable generator and discriminator backbones with configurable depth, width and scale factors.
  \item \textbf{Configuration-first workflow:} Repeatable training and evaluation using concise YAML definitions.
  \item \textbf{Training stabilization options:} Warmup, adversarial ramping, label smoothing, spectral normalization, adaptive learning-rate scheduling and optional EMA tracking.
  \item \textbf{Multispectral compatibility:} Native support for arbitrary band combinations across sensors.
  \item \textbf{OpenSR ecosystem integration:} Standardized evaluation via \texttt{opensr-test} \citep{osrtest} and scalable inference utilities via \texttt{opensr-utils} \citep{osrutils}.
  \item \textbf{HPC-ready inference:} Inference scripts for HPC-clusters enabling large-scale SR product generation.
\end{itemize}

A typical workflow in OpenSR-SRGAN involves selecting an architecture, defining the desired scale factor, specifying multispectral band ordering and choosing appropriate loss terms, all through simple YAML file modifications. The framework then handles model construction, dataset loading, training procedures and evaluation, which allows researchers to focus on experiment design rather than implementation details. This approach is particularly useful when comparing multiple configurations or running ablation studies, since each experiment is defined through its configuration file. Internally, the framework supports flexible loss composition, allowing pixel and spectral fidelity to be balanced against perceptual sharpness and adversarial realism. PyTorch Lightning's manual-optimization path was chosen to make generator and discriminator update order, pretraining, adversarial ramping, and EMA updates explicit, at the cost of not using Lightning's automatic optimizer handling. Users may also enable Wasserstein GAN training \citep{arjovsky2017wasserstein} with an optional R1 gradient penalty \citep{mescheder2018r1}. These components together form a practical and extensible foundation for SRGAN experimentation while keeping the central training trade-offs visible.

\section{Research Impact Statement}

Version 0.5.1 is distributed through the Python Package Index and archived on Zenodo. The repository provides user and API documentation, a runnable Colab example, contribution guidelines, and automated tests exercised by continuous integration on Python 3.12, 3.13, and 3.14. Resolved experiment configurations, example data, pretrained weights, and a Slurm-based large-area inference workflow provide concrete reproducibility and community-readiness evidence beyond the software source alone.

OpenSR-SRGAN is the adversarial-model component of the European Space Agency OpenSR project. It includes a loader for the published SEN2NAIP dataset \citep{sen2naip}, produces models that can be evaluated with \texttt{opensr-test} \citep{osrtest}, and integrates \texttt{opensr-utils} \citep{osrutils} for tiled inference on large Sentinel-2 products. Public pretrained configurations and checkpoints support ongoing experiments on four-band RGB--NIR and six-band short-wave-infrared super-resolution.

\section{Limitations}

Super-resolution methods enhance apparent detail but cannot replace imagery collected at native high resolution. OpenSR-SRGAN focuses on flexibility and reproducibility rather than state-of-the-art performance, and results depend on proper preprocessing and accurate LR--HR alignment. GAN training and SR in general remain sensitive to dataset size and diversity, and may produce instability or spectral artifacts, particularly when reference data are geographically skewed, biased by land cover, or limited in size and scope.

\section{Licensing and Availability}

The source code is made available through the \href{https://github.com/ESAOpenSR/SRGAN}{ESAOpenSR/OpenSR-SRGAN Github repository}. Full documentation, API references, quickstart guides and tips and tricks can be found at \url{https://srgan.opensr.eu}. A reproducible notebook is permanently hosted on \href{https://colab.research.google.com/drive/16W0FWr6py1J8P4po7JbNDMaepHUM97yL?usp=sharing}{Google Colab}.

In the spirit of open science and collaboration, we encourage feature requests and updates, bug fixes and reports, as well as general questions and concerns via direct interaction with the repository. \texttt{OpenSR-SRGAN} is licensed under the Apache-2.0 license.

\section{AI Usage Disclosure}

OpenAI Codex (versions 5.1--5.5) was used to assist with and generate parts of the documentation, automated tests and continuous-integration workflows, and model training and implementation code. The authors reviewed all AI-assisted changes and retain responsibility for the final code and functionality. Software changes are verified through targeted tests, the continuous-integration suite, and frequent manual usage. Experimental data, results, imagery, and pretrained model weights are fully reproducible.

The manuscript has been edited for clarity and readability only, using Writeful embedded in Overleaf.

\section{Acknowledgement}

This work has been supported by the European Space Agency (ESA) $\Phi$-Lab, within the framework of the \href{https://eo4society.esa.int/projects/opensr/}{``Explainable AI: Application to Trustworthy Super-Resolution (OpenSR)''} Project.

\bibliographystyle{plainnat}
\bibliography{references}

\end{document}